# Database of handwritten Arabic mathematical formulas images

Ibtissem HADJ ALI, Mohammed Ali MAHJOUB
Research unit SAGE
National Engineering School of Sousse (Eniso), University of Sousse, Tunisia
e-mail : ibtissemhadjali@gmail.com
medali.mahjoub@ipeim.rnu.tn

*Abstract*—Although publicly available, ground-truthed database have proven useful for training, evaluating, and comparing recognition systems in many domains, the availability of such database for handwritten Arabic mathematical formula recognition in particular, is currently quite poor. In this paper, we present a new public database that contains mathematical expressions available in their off-line handwritten form. Here, we describe the different steps that allowed us to acquire this database, from the creation of the mathematical expression corpora to the transcription of the collected data. Currently, the dataset contains 4 238 off-line handwritten mathematical expressions written by 66 writers and 20 300 handwritten isolated symbol images. The ground truth is also provided for the handwritten expressions as XML files with the number of symbols, and the MATHML structure.

*Keywords—Mathematical expression recognition; database; Handwritten; Arabic formula.*

I. INTRODUCTION

The systems of Handwritten text recognition have achieved recently significant progress, thanks to developments in segmentation, recognition and language models. Those systems are less powerful when the languages to be recognized have a two dimensional layout. This is the case for mathematical expressions [1]. Mathematics has a number of characteristics which distinguish it from conventional text and make it a challenging area for recognition. This include principally its two dimensional structure and the diversity of used symbols, especially in Arabic context. Note that recognition of mathematical Latin formulas has been widely studied in past years but, few are works that delve into recognition of Arabic mathematical formulas [2-3]. Recognition of mathematical formulas implies being capable of solving three sub problems: segmentation which as a result a list of connected components and their attributes (location, size, etc.), the second problem is the symbol recognition, During this step, each symbol candidate is passed to a classifier. Finally the third step is the symbol arrangement analysis which is particularly hard for mathematics, as it may be difficult to decide what the exact relation of two or more symbols is.

Many recognition domains have benefited from the creation of large, realistic corpora of ground-truthed input. Such corpora are valuable for training, evaluation, and regression testing of individual recognition systems. They also facilitate comparison between state-of-the-art recognizers. Accessible corpora enable the recognition contests which have proven useful for many fields, such as the field of recognition of Latin mathematical formula which presents a datasets that facilitates the progress of this domain like the dataset HAMEX [4] which represent a public dataset that contains mathematical expressions available in their on-line handwritten form and in their audio spoken form, also the also the ground-truthed corpora presented by S.MacLean,G.Labahn, E.Lank ,M.Marzouk and David.Tausky in [5] which provide a publicly available corpus of roughly 5,000 hand-drawn mathematical expressions on-line, these expression are transcribed by 20 different student, then automatically annotated with ground-truth. This corpus was created as a tool for training and testing the math recognition engine of MathBrush. Another freely available source of expressions is the set used by Raman for his Ph.D. work [6].There is also the database of Grain and Chaudhuri [7], it is a corpus for OCR research on printed mathematical expressions, this database formed by 400 scientific and technical document images containing mathematical expressions. For each document, its embedded and displayed expressions are collected into two different files. the field of printed and handwritten Arabic OCR systems has benefited from the availability of public data sets, such as the IFN/ENIT database [8] of Arabic handwritten words, ADAB database [9] of segmented online handwritten Arabic characters and the APTI database [10] which is a large-scale benchmark for printed text recognition.

However, to our knowledge, no attempts have yet been made on the development of data sets for Arabic handwritten mathematical formulas, despite that the Arab mathematical notation used in manuals and school curriculum in middle east countries. This obstacle to the progress of work on the recognition of Arab mathematics who is the domain of our research.

Therefore Considering this, we have initiated the development of a large database of images of Arabic handwritten mathematical formulas and the symbols that composes formulas handwritten an isolated way. This database will be used for our own research and will be made available for the scientific community to evaluate their recognition systems. The database has been named HAMF for Handwritten Arabic Mathematical Formulas and it contains scanned images of mathematical formulas transcribed by 66 students and researchers at the National School of Sousse engineer (ENISo).

The objective of this paper is to describe the HAMF database. In section II, we present details about the specificity of the Arabic mathematical notation. the handwritten acquisition process and the ground-truth is presented in section III. In section IV details about the database and its organization structure are presented. Finally some conclusions are presented in Section V.

## II. ARABIC MATHEMATICAL NOTATION OVERVIEW

In the Arabic Presentation , mathematical expressions are written right to left,  for example, -1 might be written as 1- and using Arabic symbols from its alphabet. These symbols are used to note the names of variables and unknown functions. As for the names of usual functions, abbreviations of the names of these functions are used, Table I provides some usual functions and their latin equivalents.

Arabic notation uses either the same symbols as those used in current use (eg +, -,≠.) or the same symbols through an inversion sense (ex. < and >, → and ←), or Latin symbols reflected. These symbols are images mirrors Latin symbols, such as the square root, the integral and the sum Fig. 1 gives some examples of Latin symbols reflected. Arabic notation used in different regions, two number systems either Arab or Arab-Hindu.

- Arabic numerals:  0, 1, 2, 3, 4, 5, 6, 7, 8, 9
- Arabic -Hindu numerals: ٠, ١, ٢, ٣, ٤, ٥, ٦, ٧, ٨, ٩

TABLE I. USUAL FUNCTIONS AND THEIR LATIN EQUIVALENT

| Arab | Latin |
|---|---|
| sin | جا |
| cos | جتا |
| tan | ظا |
| cot | ظتا |
| lg | لو |
| lim | نها |

## III. DESCRIPTION OF THE DATABSE

To be complete implies that a corpus contains all variations and combinations of syntax for a particular domain. To be representative implies that a corpus reflects the natural distribution of these syntactic elements in real-world usage, and it must be ground-truthed with very high accuracy. This section explores the difficulties associated with attaining these properties, and describes all the steps to create a ground-truthed database of  handwritten Arabic  mathematical expressions.

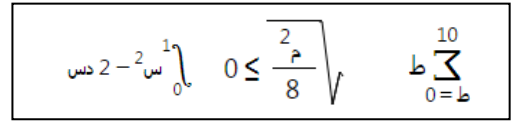

Fig. 1.  Latin symbols reflected

### A. Data collection and transcription process

Choosing the right set of data is always an important aspect of testing any system performance. In the case of mathematical expression recognition, the main difficulty in building a corpus  is to find realistic expressions from the real world. Some approaches generate such a corpus from a grammar [5], but it supposes that the grammar used is representative of the language. Thus, the best way is to use authentic data. In our case, we create a corpora composed by 65 different expressions. These expressions have different structures, layouts and geometric complexity. They  also represent the variability in terms of expression symbols because the number of symbols that constitutes a formula varies between 5 and 18 with an average of 10 symbols by formula. Table II gives details on the symbols composing the corpus vocabulary.

TABLE II. SYMBOLS COMPOSING THE CORPUS VOCABULARY

| Classes | Symbols |
|---|---|
| Arabic characters | أ ت م ح ط ق د و ع<br>ص ل ن ك(or ك) س ر ج ب |
| Digits | 0...9 |
| Operators | + - × |
| Equality op. | = ≠ ≥ > < ≤ |
| Elastic op. | ___ √ ∫ \ ∏ |
| Functions | لو نها جتا جا ظتا ظا |
| Others | ← ∞ ( ) |

To be complete and have a representative coverage expressions used in the corpus covers several mathematical areas which are mainly: arithmetic operation, equation, inequalities, limit, fraction, trigonometric functions, logarithmic and  integrals .

It was our goal to acquire a database of handwritten mathematical expressions that are all contained in the corpus that we have created. For this purpose, it was generated 5 forms named "A", "B", "C", "D", "E", and each form contains a parity of expression of the corpus to transcribe by Volunteers writers, the table III gives the number of expressions in each form.

Fig. 2. An example of a filled form

Each form consists of three parts, a filled example of the devised form is shown in Fig 2, the first block include general information about the writer : the name, age, gender, handiness and writer's profession. In the second part of the form we find a grid which contain at the right the printed expressions and at the left empty boxes where the writers have to transcribe in their handwriting the expressions samples. In the last of page we give the writer's number and the form identifier which is a combination between the form name and the writer's number, for example "A_002".

in a mathematical formula recognition engine ,among the interesting steps we find the recognition of segmented symbols. for this reason we opted in this database to transcribing the 51 isolated symbols that composes the different formulas of our corpus, to used it as a training set. Table 2 gives all the symbols to collect. For this purpose it was generated 3 forms named "Symb1", "Symb2", "Symb3", that contain the different isolated symbols and each symbol must be transcribed 8 times by the same writer. Fig 4 gives an example of a form of isolated symbols.

To create the database More than 65 people, most of them selected from the narrower range of the National Engineering School of Sousse (ENISo), contributed to the database. Each writer was asked to fill the forms with handwritten and transcribe the printed formulas and the isolated symbols.

Fig 3 shows the images of a mathematical formula transcribed by different writers.

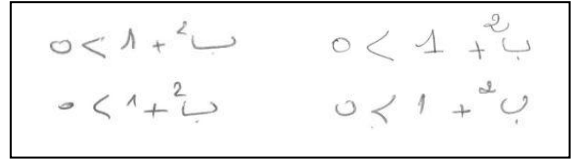

Fig. 3. Examples from the database: A mathematical formula written by four different writers.

TABLE III. NUMBER OF EXPRESSIONS IN EACH FORM.

| Form | Number |
|---|---|
| Form A | 14 |
| Form B | 12 |
| Form C | 13 |
| Form D | 13 |
| Form E | 13 |

B. *Forms processing and handwritten formulas extraction*

All form pages were scanned into TIFF files with 300 dpi and binarized automatically by the same scanner .While scanning a page the page identifier was keyed in manually. For extracting the handwritten mathematical expressions and symbols from a scanned form, a number of image preprocessing and segmentation algorithms have been developed. First, the skew of the document is corrected. Then the positions of the horizontal and vertical lines are computed by a projection method. Given this positional information and using an advanced projection method who was performed to extract the handwritten mathematical expressions and symbols images on the page automatically. The whole HAMF database consists of 4 238 images of handwritten Arabic mathematical formula and 20 300 isolated symbols images. Table IV gives examples of handwritten isolated symbols

TABLE IV. EXAMPLES OF HANDWRITTEN ISOLATED SYMBOLS

## C. Ground-truth

The benefits of recognizer training and the validity of recognition accuracy measurements are limited by the correctness of corpus ground-truth. For this reason, each expression image in this database is fully described using an XML file containing ground truth . An example of such ground-truth XML file is given in Fig 5.

The XML file is composed by three markups sections:

- "FormuleInfo": this element contains sub-elements that indicate the number of the writer, Form identifier and location number of the formula in the form.

- "Nb_Symb": in this element, we specify the number of symbols constituting the mathematical expression.

- "MathMl": in this element, we give the MATHML representation

MathML is built upon XML and is used to express formulas for mathematics and science in general. Formulas are expressed in MathML as a tree of XML tags. In the MathMl representation Symbols are classified using the <mn>, <mi> or <mo> tags for numbers, variables and operators, respectively. While 10 is inside a <mn> tag, each letter x and y are isolated in different <mi>.

An Arabic formula like $\overline{\text{س}+3}$ vould be expressed as:

<math dir="rtl">

  <msqrt>

       <mi>س</mi><mo>+</mo><mn>3</mn>

  </msqrt>

</math>

The "dir" attribute determines that the formula is presented from right to left. No other indication that a formula is Arabic would be found than the "dir" attribute and the Arabic characters by themselves [11].

Fig. 4. An example of a filled form of isolated symbols

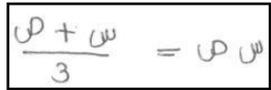

(a)

(b)

Fig. 5. (a) image of handwriting formula, (b) the xml file ground-truth of the formula in figure (a)

## IV. DATABASE DETAILS

In the total the database consists of a total of 328 filled form images that provides 4 238 handwriting Arabic mathematical expressions that contains 41 266 different symbols written by 66 writers, and 150 other filled forms for the isolated symbols which added to the base 20 300 handwritten isolated symbols.

The general structure of the database is shown in Fig 6. The database contains two partition. The first partition includes the forms and images of handwritten isolated symbols, the second is divided into five set: set_A, set_B, set_C, set_D, set_E, to allow for flexibility in the composition of development, training and testing partitions . All the five sets share the same structure. Every series contains three folders. the first is the Forms folder which contains the images of the original data-entry forms that have been used to collect the samples in grayscale versions, every set contains forms that includes the same type of expressions. The name for each form is the combination of the letter that designates the set name and number of the writer e.g. "B_020.tif". the second folder includes the image of the handwritten mathematical formula, each image name introduced the name of the form which is extracted and the position number of the formula in the form e.g. " B_020_10.tif". For each formula image is corresponding an xml ground truth file, which has the same name of the formula image. All the ground truth file of any set are saved in folder named "Truth". Table 5 provides an overview of the number of expressions and symbols by set.

The HAMF database of handwritten Arabic mathematical formulas images is publicly available for the purpose of research. It can be ordered by sending an e-mail to one of the authors.

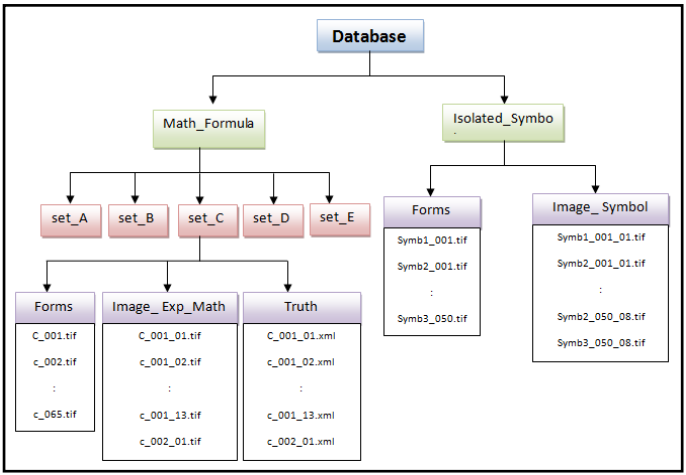

Fig. 6. The general structure of the Database

TABLE V. QUANTITY OF EXPRESSION AND SYMBOLS IN THE DATABASE BY SET

| Set name | Quantity of expressions | Quantity of symbols |
|---|---|---|
| set_A | 910 | 7 800 |
| set_B | 777 | 8 320 |
| set_C | 850 | 8 712 |
| set_D | 848 | 8 382 |
| set_E | 853 | 8 052 |
| Total | 4 238 | 41 266 |

## V. Conclusion

In this paper, we presented HAMF, a new database. This database is freely available and contains about 4 238 off-line handwriting Arabic mathematical expressions written by 66 different writer and 20 300 off-line handwriting isolated symbols that composites the mathematical formulas . We have shown how this dataset has been drawn up, from the choice of the mathematical expression corpora to the transcription of the collected data. At the end, the handwritten mathematical expressions are provided with their ground truth.

This database will help in bridging the evaluation gap between diverse systems dedicated to recognizing off-line handwritten Arabic mathematical expressions that use their own database.